\documentclass[twocolumn,3p]{elsarticle}

\usepackage{hyperref, lineno}

\journal{Pattern Recognition Letters}

\usepackage{amsmath,amssymb,amsfonts}
\usepackage{algorithm, algpseudocode}
\usepackage{graphicx}
\usepackage{textcomp}
\usepackage{xcolor}
\def\BibTeX{{\rm B\kern-.05em{\sc i\kern-.025em b}\kern-.08em
    T\kern-.1667em\lower.7ex\hbox{E}\kern-.125emX}}
    
\newcommand{\ones}{{\mathbf 1}}   
\newcommand{\x}{\mathbf{x}}

\newcommand{\z}{\mathbf{z}}

\newcommand{\y}{\mathbf{y}}
\newcommand{\kfun}{\mathbf{\kappa}}
\newcommand{\bfTheta}{\boldsymbol\Theta}
\newcommand{\X}{\mathbf{X}}
\newcommand{\F}{\mathbf{F}}
\newcommand{\B}{\mathbf{B}}
\newcommand{\bfH}{\mathbf{H}}
\newcommand{\K}{\mathbf{K}}
\newcommand{\I}{\mathbf{I}}
\newcommand{\reg}{\mathrm{reg}}
\newcommand{\ks}{\mathbf{k}}
\newcommand{\tr}{\mathrm{train}}
\newcommand{\bfeps}{\boldsymbol\epsilon}
\newcommand{\zero}{\mathbf{0}}

\newcommand{\target}{\mathrm{target}}
\newcommand{\bfOmega}{\boldsymbol\Omega}
\newcommand\numberthis{\addtocounter{equation}{1}\tag{\theequation}}

\usepackage{amsmath}

\begin{document}

\begin{frontmatter}

\title{Uncertainty quantification for multiclass data description}

\author[mymainaddress]{Leila Kalantari\corref{mycorrespondingauthor}}
\ead{leila@ufl.edu}
\cortext[mycorrespondingauthor]{Corresponding author}

\author[mymainaddress]{Jose Principe}
\ead{principe@cnel.ufl.edu}

\author[mysecondaryaddress]{Kathryn E. Sieving}
\ead{chucao@ufl.edu}

\address[mymainaddress]{Electrical \& Computer Engineering, University of Florida}
\address[mysecondaryaddress]{Wildlife Ecology \& Conservation, University of Florida}

\begin{abstract}
In this manuscript, we propose a multiclass data description model based on kernel Mahalanobis distance (MDD-KM) with self-adapting hyperparameter setting. MDD-KM provides uncertainty quantification and can be deployed to build classification systems for the realistic scenario where out-of-distribution (OOD) samples are present among the test data.  Given a test signal, a quantity related to empirical kernel Mahalanobis distance between the signal and each of the training classes is computed. Since these quantities correspond to the same reproducing kernel Hilbert space, they are commensurable and hence can be readily treated as classification scores without further application of fusion techniques.  To set kernel parameters, we exploit the fact that predictive variance according to a Gaussian process (GP) is empirical kernel Mahalanobis distance when a centralized kernel is used, and propose to use GP's negative likelihood function as the cost function.  We conduct experiments on the real problem of avian note classification. We report a prototypical classification system based on a hierarchical linear dynamical system with MDD-KM as a component.  Our classification system does not require sound event detection as a preprocessing step, and is able to find instances of training avian notes with varying length among OOD samples (corresponding to unknown notes of disinterest) in the test audio clip.   Domain knowledge is leveraged to make crisp decisions from raw classification scores. We demonstrate the superior performance of MDD-KM over possibilistic $K$-nearest neighbor.\end{abstract}


\end{frontmatter}


\section{Introduction}

Assuming a classifier is only applied to signals from the training classes at the test time limits the range of real-world applications where pattern recognition can have an impact. Classification in the presence of out-of-distribution (OOD) samples at the test time is prevalent in many scientific and engineering problems. For example, in environmental bioacoustics, detection or classification of target sounds from remotely recorded audio data facilitates numerous objectives, from species inventory to locating illegal logging and poaching activities in real time
\cite{bardeli2010detecting,potamitis2014automatic}.
Soft (as opposed to crisp) memberships are desired due to the fact that being OOD is relative, and also the fact that the training classes may overlap. Instead of making a hasty crisp decision for a given test signal, soft memberships, aka.~class scores, could be propagated to other components of a classification system (such as expert knowledge and/or classifiers based on other features) in order to resolve some of the uncertainties.
This is in accordance to David Marr's
principle of least commitment, which states decision making should be postponed as
long as possible \cite{marr1982vision}.  Therefore, classifiers that can properly quantify uncertainty in multiclass settings need to be developed.   

Probabilistic and deep learning classification methods, and in general supervised classification methods do not reliably quantify (membership) uncertainty on OOD samples \cite{szegedy2013intriguing, fawzi2015fundamental}.  Probabilistic classifiers such as Gaussian process classification (GPC) and support vector machine with Platt scaling (SVM-Platt) are derived under the assumption that the probabilities of a test signal belonging to the training classes 
must sum to one.  This is the algebraic equivalence of a test signal belonging to one/some of the training classes.  Such classifiers cannot distinguish between OOD and ambiguous samples (occurring in class overlap).  In a two-class scenario, both samples are mapped to probability $0.5$.   Possibilistic $K$-nearest neighbors (PKNN) was developed to address this shortcoming  \cite{frigui2009detection}.  A PKNN's possibility represents the resemblance of the point to the class, regardless of its resemblance to other classes. Naturally, PKNN does not readily distinguish among training classes as well as probabilistic classifiers do because it does not leverage the sum-to-one constraint, and is not trained under a cost function based on classification error.  However, it outputs proper class scores to be inputted to other components of a classification system and the whole system has superior distinguishability power in addition to being superior in discerning outliers.  PKNN was applied successfully to the problem of landmine detection and is still operational in the field.  In this paper, we present MDD-KM as another approach to uncertainty quantification for multiclass memberships besides PKNN.  The two are empirically compared.

PKNN starts with clustering the training data using self-organizing maps.  This is for two reasons: 1) addressing class imbalance and 2) robustness (wrt.~outliers in the training data) MDD-KM, as proposed is this paper, is not robust with respect to the outliers in the training data since it is based on sample covariance which is known to be sensitive to outliers (see chapter 8 of \cite{huber2004robust}).  In our prototypical classification system, prior to applying MDD-KM, the training data from recordings is processed using a hierarchical linear dynamical system (HLDS) model for feature extraction and mitigating the effect of outliers.  HLDS, which is a general time series representation model, has been adapted and shown to work well for this application \cite{kalantari2019hierarchical}.  MDD-KM as of now is suitable in these situations:  1) The training data is already treated for outliers, and 2) deep features are not needed.  
To be able to apply MDD-KM in settings that violate these two conditions, is a subject of future research.  Note that to be able to deploy MDD-KM, no outlier detection is necessary as a preprocessing step to test.  However, outliers in the training data have to be mitigated. There is a difference between the treatment of outliers in these two settings.  For treatment of outliers in the training data, one should see literature on robust optimization and empirical risk minimization.  The treatment of outliers at the test time is related to adversarial robustness and OOD detection.


We chose avian note classification as our example application for several reasons.  1) The problem fits the criteria (presence of OOD at the test time). 2) It works with the limitation of MDD-KM in not being robust since we already know how to mitigate the effects of outliers in training data using HLDS. 3) With this application, the ability of MDD-KM (as a component in a classification system) in performing one-shot segmentation and classification can be demonstrated.  No sound event detection is needed as a preprocessing step during the test.  This is specially important for this application since there is 
 a mismatch between the training data (recordings of individual notes) and the test data (long recordings containing many notes).  4) With this application, the ability of MDD-KM in dealing with varying length note instances can be demonstrated.  Different instances of the same note can be of different lengths.  Our system does not use zero padding or other techniques to force instances to be of equal lengths.

The paper is organized as follows.  Section \ref{sec_back} is on technical background.  Section \ref{sec_method} describes our method and its novelty.  Section \ref{sec_exp} describes our prototypical system for avian note detection, and includes quantitative comparisons with PKNN.  Section \ref{sec_future} is about the future research.

\section{Technical background}
\label{sec_back}
Section \ref{sec_hlds} describes HLDS.   It is the first component of the prototypical system we devised for avian note classification (Section \ref{sec_avian}).   Section \ref{sec_occkmd} MDD-KM in one-class setting and its relationship to Gaussian process (GP).  Possibilistic $K$-nearest neighbor (PKNN) is described in Section \ref{sec_pknn}.  PKNN is the soft classification method with OOD detection that we deemed best to compare against MDD-KM in Section \ref{sec_result}.

\subsection{HLDS}
\label{sec_hlds} 
HLDS is an architecture for linear modeling of time series data proposed by \cite{cinar2018hierarchical}.  Later, we adapted HLDS for multiscale time series representation \cite{kalantari2019hierarchical}. 
The adapted HLDS is suitable to represent time domain structure directly from the audio recording containing avian notes  before applying MDD-KM for two reasons: 1) It is label-agnostic hence suitable for representing OOD samples as well as samples from the training classes 2) It is a state model, i.e.,~it exploits the fact that observations are correlated in time to be able to smooth the data.  

We describe HLDS mathematically here.  For the ease of exposition, assume $L=2$, where $L$ is the number of hidden layers.  Let $M$ be the dimension of observation vector $\y_t$, and $N$ and $S$ be the dimensions of states $\x_t$ and $\z_t$ respectively such that $N$ is divisible by $S$.  
HLDS is described by the following system of equations to model time series $\{\y_t\}$:
\begin{align}
\z_t &= \z_{t-1}+\bfeps^\z_t \label{eq_z} & \mbox{2nd hidden layer}\\
\x_t &= \x_{t-1} + \B  \z_{t-1} + \bfeps^\x_t \label{eq_x} & \mbox{1st hidden layer}\\
\y_t &= \bfH \x_t      + \bfeps^\y_t \label{eq_y} & \mbox{observation layer}
\end{align}
where vector $\bfeps^\z_t \sim \mathcal{N}(\zero,r^\z\I_S)$ is the innovation for the last hidden layer, vector $\bfeps^\x_t \sim \mathcal{N}(\zero,r^\x\I_N)$ is the
 innovation for the first hidden layer, vector $\bfeps^\y_t \sim  \mathcal{N}(\zero,r^\y\I_M)$ is the measurement error, $\bfH$ is the observation matrix, and $r^\z\I_S, r^\x\I_N, r^\y\I_M$ are covariance matrices for the corresponding Gaussian distributions. Coupling matrix $\B$ for the first hidden layer (and in general for the $l$th hidden layer st.~$1 \leq l < L$), is described by
$\B =  \left( \begin{matrix} \B_1, \dots, \B_S						 
    \end{matrix}\right)^T$. For each $s \in \{1, \dots, S\}$, $\B_s$ is an $N/S \times S$ matrix whose $s$th column is $(2S/N)  \ones$, and is zero elsewhere. 
To be able to solve the model using Kalman filtering, equations (\ref{eq_z})-(\ref{eq_y}) can be rewritten as:
\begin{align}
\tilde{\x}_t &= \tilde{\F}\ \tilde{\x}_{t-1}+ \tilde{\bfeps}_t \label{eq_joint1} & \mbox{ \ \  \hspace{-1cm}Augmented hidden layer}\\
          \y_t &= \tilde{\bfH}\ \tilde{\x}_t + \bfeps^\y_t \label{eq_joint2} & \mbox{ \ \ \hspace{-1cm} Augmented observation layer}
\end{align}
where
  $\tilde{\x}_t =   \left(  \begin{matrix} \z_t \\ \x_t \end{matrix} \right)$ is the augmented hidden state, 
  $\tilde{\F} = \left(  \begin{matrix} \I & \zero \\ \B & \I \end{matrix} \right)$ is the augmented transition matrix,  $\tilde{\bfH} = \left(\zero \ \ \bfH\right)$  is the augmented observation matrix, and vector $\tilde{\bfeps}_t = \left(  \begin{matrix} \bfeps^\z_t\\ \bfeps^\x_t  \end{matrix} \right)$ is the augmented innovation vector.   Kalman filtering solves the model efficiently in an online fashion. 


\subsection{One-class classification with Gaussian process (OCC-GP)}
\label{sec_occkmd}
One approach to uncertainty quantification in multiclass settings is to train an ensemble of one-class models.  
 MDD-KM's training results in an ensemble of OCC-GP components that share the same hyperparameters (in contrast to an ensemble of independently trained one-class models).
OCC is a challenging problem and despite recent advances in deep learning, OCC has benefitted little.  The main challenge with OCC is that information on outlier classes cannot be used during the training.  Hence, a performance-based cost function cannot be used to find parameters (such as kernel parameters).  For example, a well-tuned one-class model would maximize both recall and precision, but when only focal class samples are available (ie.~OCC setting), precision cannot be estimated. With the same reasoning, false positive rate (FPR) cannot be estimated either.  Therefore, a model which balances low FPR against high recall cannot be trained. 
Note that the challenge of multiclass classification with the presence of OOD samples at the test time, is similar to that of OCC.  In both settings, information on OOD samples should not be used for training.


OCC-GP, as we proposed \cite{kalantari2018patternRecognition}, is not based on a performance-based cost function to tune hyperparameters.  Hyperparameters are tuned by learning a low-degree polynomial (such as $y_\target(\x) \triangleq \x^T \x$) on $\x_1, \dots, \x_N$, the samples of the focal class $\bfOmega_c$.  We refer to $y_\target$ by target function in this manuscript. The cost function is  $g(\bfTheta) \triangleq \y^T \K_\reg^{-1} \y + \log | \K_\reg |$ where $\y\triangleq \left(y_\target(\x_1),\dots, y_\target(\x_N)\right)^T$ and $\K_\reg \triangleq \left( \kfun_\reg(\x_m,\x_n)\right)_{m,n}$.  Such a smooth target function definition together with the proposed cost function would promote parameter settings where nearby signals are assigned similar values.  Uncertainty score is calculated as:  $d_c(\x_*) \triangleq k_{**} - \ks_*^T \K_\reg^{-1} \ks_*$.  
Hence, Algorithm \ref{alg_main}, which describes MDD-KM, can be used for OCC-GP as well by setting the the total number of training classes to be one ($C=1$).  Note that $g(\bfTheta)$ is GP's likelihood function and $d_c(\x_*)$ is predictive variance according to GP.  It was proven that GP's predictive variance is kernel Mahalanobis distance when centralized kernel is used \cite{pekalska2009kernel}.

\subsection{PKNN}
\label{sec_pknn}
MDD-KM will be compared with PKNN. 
Frigui et al.~proposed PKNN to softly detect OOD samples as well as ambiguous data for the purpose of land mine detection using data from ground penetrating radar \cite{frigui2009detection}.  To overcome the problem of class imbalance (landmine vs.~clutter) as well as mitigating the effects of outliers, they used self-organizing maps (SOMs) to
find prototypes for each class during the training phase, where the number of prototypes are fixed across the classes.
 At test time, the similarities of a given test signal to
its $K$-nearest prototypes (obtained during training) is calculated. Using this similarity and the representativeness of each prototype to its class, a possibility (between zero and one) is calculated.   These possibilities do not have to sum to one.  If all possibilities are low, the test signal is considered to be unknown/outlier. If more than one possibility is large, the signal is considered to be ambiguous; it can belong to any of the corresponding classes. If only one of the possibilities is large the signal is declared to belong to the corresponding class with high certainty.  
\section{The proposed method}
\label{sec_method}
The training of MDD-KM is presented in Section \ref{sec_train}.  The testing is described in Section 
\ref{sec_test}.  Section \ref{sec_novelty} is about the novelty of our approach.

\subsection{Training}
\label{sec_train}
Training in our context refers to learning the kernel and the regularization parameters.   
The squared exponential (SE) is used as the kernel. SE kernel is defined as:
 \begin{equation}
 \label{eq_kernel}
 \kfun(\x_m, \x_n) \triangleq \sigma^2 \exp(- ||\x_m- \x_n ||^2/\ell^2)
  \end{equation}
 where $\sigma$ is the signal variance parameter and $\ell$ is the length-scale parameter.  The regularized kernel is defined as:
 $$
 \kfun_\reg(\x_m, \x_n) \triangleq \kfun(\x_m, \x_n) + \sigma_\reg^2
 $$

Let $\x_1, \dots, \x_N$ be the training signals from all classes and $\bfTheta$
denote the set of all parameters of MDD-KM, and $g(\bfTheta)$ denote the cost function used in the training of MDD-KM.  We defined the cost function as:
\begin{align}
\label{eq_cost}
g(\bfTheta) \triangleq \y^T \K_\reg^{-1} \y + \log | \K_\reg |
\end{align}
subject to
\begin{align}
\label{eq_target}
y_\target(\x) \triangleq \x^T \x 
\end{align}
where
 \begin{align}
 \y & \triangleq (y_\target(\x_1), \dots, y_\target(\x_N) )^T\\
 \K_\reg &\triangleq \left( \kfun_\reg(\x_m,\x_n)\right)_{m,n}
 \end{align}

 Such a smooth target function together with the cost function would promote parameter setting where nearby signals are assigned similar values according to the prediction function described by
 Equation (\ref{eq_predFun}).

The pseudocode of MDD-KM is described in Algorithm \ref{alg_main}. The cost function, Equation (\ref{eq_cost}), is expressed at line 6 of the pseudocode, with Equation (\ref{eq_target}) as constraint on line 4 of the pseudocode. 

 \begin{algorithm}[htbp]
\begin{algorithmic}[1]
\Procedure{MDD-KM}{$\x_*, \{\X_c \}_{c=1}^C; \ \kfun, y_\target$}
   \State $\X_\tr \triangleq \left(\X_1, \dots, \X_c\right)$
  \State $\kfun_\reg(\x,\x) \triangleq \kfun(\x,\x) +\sigma_\reg^2$\Comment{Regularized kernel function}  
           \State $\y \triangleq y_\target(\X_\tr)$
   \State $\K_\reg \triangleq \kfun_\reg(\X_\tr,\X_\tr)$
   \State $\hat{\bfTheta} \gets \min_{\bfTheta} \left(\y^T \K_\reg^{-1} \y + \log |  \K_\reg  | \right)$\Comment{Parameter learning}
    \ForAll{$c \in \{1,\dots,C\}$}
    \State $d_c(\x_*) \triangleq \kfun(\x_*,\x_*)- \kfun(\X_c,\x_*)^T \kfun_\reg(\X_c, \X_c)^{-1} \kfun(\X_c,\x_*)$\Comment{Class score for $\x_*$ (membership to $\bfOmega_c$)}
    \EndFor
   \State \Return $\{d_c(\x_*)\}_c$
\EndProcedure   
\end{algorithmic}
\caption{MDD-KM for a fixed target function $y_\target$ and kernel form $\kfun$ (with unknown parameters).  Training is finding hyperparameters $\bfTheta$ (kernel and regularization parameters)  and is described at line 6. Class-wise scores for a given test signal $\x_*$ are computed at line 8.
When $\kfun \triangleq  \sigma^2 \exp(- ||\x_m- \x_n ||^2/\ell^2)$, function $y_\target$ can be defined to be a low degree polynomial such as $y_\target(\x) \triangleq \x^T \x$. Scalar $\sigma_\reg$ should be the smallest value that makes matrix $\kfun(\X_\tr,\X_\tr)$ well-conditioned. Columns of matrix $\X_c$ are the training samples from class $\bfOmega_c$.}
 \label{alg_main}
\end{algorithm} 

\subsection{Testing}
\label{sec_test}
The membership uncertainty to class $\bfOmega_c$ for signal $\x_*$ is algebraically expressed as:
\begin{align*}
 \numberthis\label{eq_predFun}
& d_c(\x_*) \\
 &\triangleq \kfun(\x_*,\x_*)- \kfun(\X_c,\x_*)^T \kfun_\reg(\X_c, \X_c)^{-1} \kfun(\X_c,\x_*)
\end{align*}
which is also described at line 8 of Algorithm \ref{alg_main}.  Note that the signals used in training (line 6) and testing (line 8) are different.  Training signals from all classes ($\X_\tr$) are used when learning parameters.  During test, only focal class signals ($\X_c$) are used when calculating class scores for a given test signal and class $\bfOmega_c$.

\subsection{Novelty}
\label{sec_novelty}
MDD-KM is a generalization of OCC-GP where the number of training classes can be more than one.  It is similar to OCC-GP
 with respect to the restriction that information on OOD samples cannot be used for the training and still at the test time the model is expected to discern OOD samples from the samples of the training class(es) (Section \ref{sec_occkmd}). Here, we summarize how MDD-KM is different from OCC-GP \cite{kalantari2018patternRecognition}.
\paragraph{Commensurability} As discussed in Section \ref{sec_occkmd}, one approach to uncertainty quantification in multiclass setting is to train an ensemble of independent one-class classifiers.  A shortcoming with this approach is that scores are not commensurable across classes, because they are of different scales. Hence, further application of fusion techniques are required for commensurability. In contrast, MDD-KM's raw class scores are readily commensurable because they are calculated in the {\it same} RKHS.
 The one-class version (in preparation \cite{kalantari2018patternRecognition}) and its variant \cite{kalantari2016one} did not produce commensurable outputs.
 The only previous effort at commensurability was a rank-based approach to normalize outputs in Kalantari's dissertation \cite{leila2017thesis}.  A rank-based normalization approach requires a lot of data.  Specially low classification scores are unreliable because they correspond to regions with few training samples.  As the number of training classes increases, the peril with a rank-based approach increases; it only takes a single badly normalized one-class component to overpower the rest. 
\paragraph{Experiments} Experiments in \cite{kalantari2016one}  and \cite{kalantari2018patternRecognition} are in one-class setting (measured by area under the ROC curve).  Accordingly, OCC-GP is only compared to other one-class classifiers: support vector data description (SVDD), autoencoder, and deep SVDD. 
Here, we utilize MDD-KM on a multiclass problem, and compare it with possibilistic KNN.  We also demonstrate a prototypical system where domain knowledge is leveraged to make crisp decision from class scores to demonstrate how soft scores can be useful.
\paragraph{Self-adapting parameters}
Parameters of MDD-KM are that of kernel and regularization.  They are all self-adapting in an unsupervised fashion.  They are set differently than that of our previous published work \cite{kalantari2016one}.  For $C=1$ case, the work was presented only in Kalantari's dissertation \cite{leila2017thesis}.

\section{Experiments}
\label{sec_exp}
To our knowledge there is no metric that can evaluate raw uncertainty outputs from a multiclass (soft) classification algorithm (for one-class there is ROC curve for example).  Therefore, the uncertainty quantification of MDD-KM is being evaluated within a classification system rather than a standalone algorithm.  This way the crisp final decisions of the classification system can be a measure of how useful the uncertainty outputs of MDD-KM are in practice.  Before discussing the results in Section \ref{sec_result} (and compare it with when MDD-KM component is replaced by PKNN), we describe the dataset (Section \ref{sec_avian_data}) and the system we devised where MDD-KM's raw outputs are inputted into a 
simple decision making algorithm which leverages simple domain knowledge (Section \ref{sec_avian}).

\subsection{Avian dataset}
\label{sec_avian_data}
The dataset of avian notes is described here.
 Carolina chickadee ({\it Poecile carolinensis}) and tufted titmouse ({\it Baeolophus bicolor}) are well-known for their exceptional vocal complexity, including rudimentary syntax, that serves as a model for the evolution of human language. The two species are abbreviated by 
 {\tt cach} and {\tt tuti} respectively throughout this section. Their vocal outputs encode situationally specific information about food, predators, and habitat quality of potential use in monitoring of environmental health under climate change.
 Various notes uttered by {\tt cach} and {\tt tuti} were recorded at $1$ m distance from the source, and are nearly noiseless.  The available labeled notes are
{\tt tuti-D}, {\tt tuti-F}, {\tt cach-Z}, {\tt tuti-A}, {\tt tuti-Z}, {\tt cach-C}, {\tt cach-D} and {\tt cach-E}.  The number of instances of each note is $15, 6, 4, 4, 4, 6, 7$ and $10$ respectively with a total of $56$. The sampling rate is $22050$.

\subsection{A system for segmenting and classifying avian notes from recorded audio}
\label{sec_avian}
Here, we describe a prototypical classification system in which MDD-KM is a component.  
The system has three components: 1) feature representation (using HLDS), 2) uncertainty quantification and, 3) decision making.  MDD-KM soft raw scores are used to make final crisp decisions by leveraging domain knowledge.  The described system finds instances of three avian notes from recorded audio (which includes OOD notes as well as instances from the three training notes).  Here are descriptions of system components:

\paragraph*{Feature representation} We used HLDS for feature representation for reasons discussed in Section \ref{sec_hlds}.
Time domain sliding windows $\{\y^\prime_t\}$ (from either train or test audio clips) are preprocessed as $\y_t = |A \y^\prime_t|$ before applying HLDS, where $A$ is the discrete cosine transform.  Matrix $\bfH$ is set to identity.
Let $w^\prime$ and $w$ denote the dimensions of $\y^\prime$ and $\y$ respectively. The hyperparameters of HLDS are set as follows:  the number of hidden layers is three ($L=3$), the dimensions of the hidden layers are set to $[96, 24, 12]$ which implies $w^\prime=w=96$ since $\bfH$ is set to identity; $q=48$ where $q$ denotes the number of overlapping samples between two consecutive time domain sliding windows.
Each of the (observation error and innovation) parameters $r^{\y}, r^{\x}, \dots,  r^{\z}$ is set proportional to the dimension of its corresponding layer.  $\z$-representations (one $\z_t$ per $\y^\prime_t$) are inputs to MDD-KM's training or testing steps.  Note that these parameters are for HLDS not that of MDD-KM or PKNN.

\paragraph*{Uncertainty quantification}  We assign three class scores according to MDD-KM (or PKNN for comparison) to each $\y_t$.  Each column of Figure \ref{fig_birdExample}-a is a visualization of a single $\y_t$.
Each of the three scores represents $\y_t$'s degree of membership to the corresponding training class (Figure \ref{fig_birdExample}-b).   All MDD-KM's uncertainty scores are transformed by $-\log(\sqrt{x})$. This transformation is order preserving (reversely) and is for visualization purposes.

\paragraph*{Decision making}
This last component is a simple example of how context or domain knowledge can be leveraged to make crisp decisions from soft class-wise scores. An example of crisp decisions are illustrated in Figure \ref{fig_birdExample}-c.
Making crisp decisions from soft scores of MDD-KM (Figure \ref{fig_birdExample}-b), involves several steps: 1)  All scores below threshold $\tau$ were discarded (set to $0$).  2) All non-zero scores that lasted less than $35$ consecutive time steps were discarded since a note cannot last that short.   3) A class score that is larger than other class scores for longer than $60$  time steps is declared as final crisp decision.  4) For the remaining undecided subsets, for each class, the mean (over the subset) class score is assigned to the whole segment.  5) Crisp decision for each time step is taken according to the max score.  6) Crisp decisions which last less than $35$ time steps are removed (as with step 2), and those segments are declared as OOD. 
Parameter $\tau$  was set as follows.  For MDD-KM, $\tau \triangleq 1.8/\mu$ where $\mu$ is the maximum test on train score.  For PKNN, $\tau \triangleq .0015$.  The described parameters are for the decision making step not that of MDD-KM or PKNN.

Note that this systems does not require a segmentation or a novelty detection algorithm as a preprocessing step. 
All notes are found with delay.

\subsection{Results} \label{sec_result}
\begin{figure*}[ht!]
\begin{tabular}{c}
\includegraphics[scale=0.5]{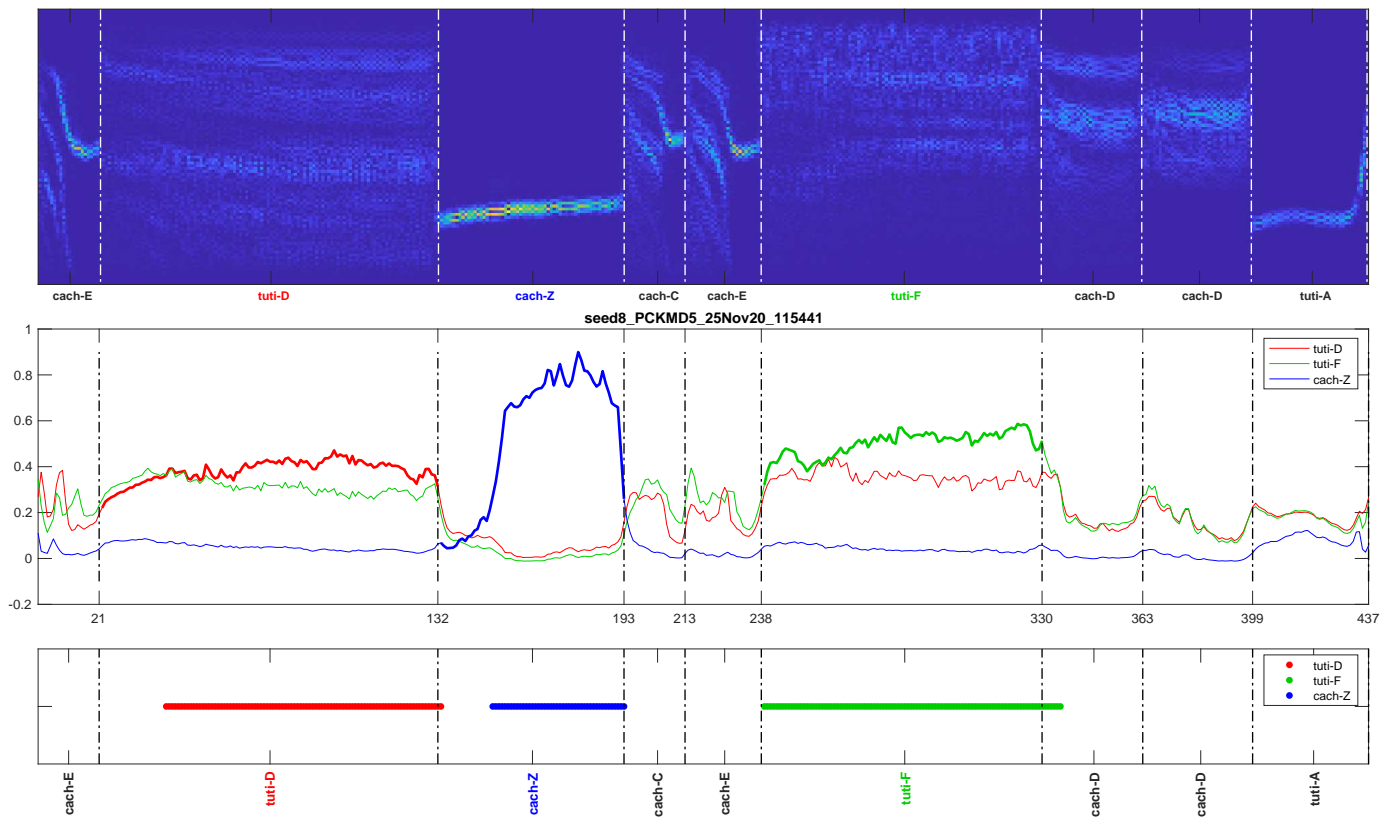}
\end{tabular}
\caption{An illustration of our classification system tested on nine unseen avian note instances (from total of $50$); {\tt tuti-D}: $3$ instances, {\tt tuti-F}: $2$ instance, {\tt cach-Z}: $1$ instance, and outliers: $5$ instances.  (a) The spectrum of the clip consisting of the nine instances.  (b) Raw scores of each time step belonging to each of the three training notes according to MDD-KM; the bolded scores are ideally above other scores in their corresponding segment (c) Final prediction for each time step.  No outlier is assigned to any of the training notes as desired.  This figure is best viewed in color.}
\label{fig_birdExample}
\end{figure*}

The training clip consists of two randomly selected instances of each of the {\tt tuti-D}, {\tt tuti-F}, and {\tt cach-Z} notes (Section \ref{sec_avian_data}).
The test clip consists of the rest of the $50$ labeled notes right after each other.  Our prototypical classification system is illustrated on a subset of the test clip in Figure \ref{fig_birdExample}.   Confusion matrix for one of the experiments (random seed 1) at the level of note instances is reported in Table \ref{tbl_confMat_avian}.  We consider all {\tt tuti-A},  {\tt tuti-Z},  {\tt cach-C},  {\tt cach-D}, and {\tt cach-E} instances as OOD.

We repeat the experiment for $50$ random seeds. For each seed, the training instances are picked randomly, and the rest of the instances are put together in the test clip in a random order.  The evaluation is performed in two ways. First, 
to include the assessment of the segmentation as well as classification, the evaluation is performed on the assignment of the sliding windows (each column of Figure \ref{fig_birdExample}-a is a visualization of a sliding window), rather than the whole note segment.  This way when a note is found with delay or the note assignment ends before the instance ends, the performance gets penalized. Macro F-scores is used for comparison to account for class imbalance. Second, the whole note segment is used as the evaluation unit.
The two evaluation results are reported in Table \ref{tbl_fscore_1}.

\begin{table}
 \begin{tabular}{ccccc}
 \hline
               &{ tuti-D}&{ tuti-F}&{ cach-Z}&OOD\\
            \hline    
{tuti-D}&12/13&0&0&1\\
{tuti-F}&0&4/4&0&0\\
{cach-Z}&0 &0 &2/2 &0\\
\hline
{tuti-A} & 0 &0 &0 &4/4\\
{tuti-Z}& 0 & 0 & 0 & 4/4\\
{cach-C}& 0 &0 &0 &6/6\\
{cach-D}&0 & 0 & 0 & 7/7\\
{cach-E}& 0 &0 &0 &10/10\\
\hline
 \end{tabular}
 \caption{Test confusion matrix for our system tested with $50$ unseen avian notes (random seed $1$).  The system was trained using six instances (two per training note).  The row titles are the true labels and the column titles are the predictions. The system correctly detects {\tt tuti-A}, {\tt tuti-Z}, {\tt cach-C}, {\tt cach-D} and {\tt catch-E} as OOD as they do not belong to any of the three training classes.}
 \label{tbl_confMat_avian}
 \end{table}
\renewcommand{\sc}{\small} 
\begin{table}
\begin{tabular}{ccccc|c}
& \sc{{\tt tuti-D}} & \sc{{\tt tuti-F}} & \sc{{\tt cach-Z}} & \sc{OOD} & \sc{overall}\\
\hline
\sc{{\it sliding}}& & & &  & \\
\vspace{-.5cm}\\
\sc{{\it window}}& & &  & &\\
\sc{MDD-KM} & .88 & .77 & .89 & .89 & \bf{.85}\\
\sc{PKNN}     & .84 & .73 & .91 & .83 & .83\\
\hline
\sc{{\it note}}& & &  & \\
\sc{MDD-KM} & .92 & .87 & .95 & & \bf{.91}\\
\sc{PKNN} &     .89 & .86 & .96 & & .90
\end{tabular}
\caption{Mean performance (F-score) over $50$ random seeds are reported for each algorithm and test unit. Once the performance is measured on classifying sliding windows (upper table) and another time on segmenting and classifying whole note instances (lower table).  Also overall performance is measured 
by macro-averaging F-scores over categories.
According to macro-average F-score, MDD-KM outperforms PKNN with $p$-value = $.003$ when the evaluation unit is a sliding window, and with $p$-value $=.386$ when the evaluation unit is a note instance (without note segmentation as a preprocessing step).}
\label{tbl_fscore_1}
\end{table}

These are notable in our setup: 1) Different instances of the same note can be of varying length.  2) In the test clip, different note instances do not occur in isolation (but not overlapping) without inserting silence or noise in between (which makes segmentation particularly challenging). 3) The system does not have sound event detection (which is a segmentation problem) or OOD detection as a preprocessing step.

\section{Future work}
\label{sec_future}
This research can be carried out at different levels of mathematical rigor or practicality.  There are many problems in many different fields that MDD-KM can potentially contribute to.  On the application level, to go beyond our prototypical system for avian notes classification and to aide answering a biologically relevant question, we will focus on automatically gathering a nearly pure set of the two species vocalizations from ambient recordings collected at 70 point locations across three states.  We will also apply MDD-KM to the problem of tree species detection using hyperspectral and lidar data since we have already applied a variant of its one-class model to this problem \cite{kalantari2016one}.  Several other applications are:  1) Freezing of gate (FoG) detection for Parkinson patients.  2) Landmine detection, which is a defense application \cite{frigui2009detection}, and 3) Malware detection, a problem in cybersecurity. It was demonstrated that supervised classification models are inadequate due to the fact that the behavior of benign programs are highly varied such that they are always undersampled \cite{miao2016malware}.  However, for some problems such as FoG deep features may be needed before applying MDD-KM. 

Matrix inversion in MDD-KM is prohibitive to its application to big data.  Therefore, another direction of future research is sparsification of MDD-KM.  Studying the statistical robustness properties of MDD-KM, in the sense formalized by Huber \cite{huber2004robust}, is a relevant theoretical question.

\bibliography{2021_arxiv_PCKMD}

\end{document}